\pdfoutput=1
\pdfoutput=1
\pdfoutput=1
\pdfoutput=1
\pdfoutput=1

\documentclass[10pt,journal,compsoc]{IEEEtran}
\ifCLASSOPTIONcompsoc
  \usepackage[compress]{cite}
\else
  \usepackage[compress]{cite}
\fi
\ifCLASSINFOpdf
   \usepackage[pdftex]{graphicx}
\else
\fi
\usepackage{times}
\usepackage{epsfig}
\usepackage{graphicx}
\usepackage{amsmath}
\usepackage{amssymb}
\usepackage{subfigure}
\usepackage{url}
\usepackage{multirow}
\usepackage{flushend}
\usepackage{color}
\usepackage{color}
\definecolor{green}{rgb}{0, 0.5, 0}
\definecolor{orange}{rgb}{0.8, 0.6, 0.2}
\definecolor{red}{rgb}{1.0, 0.0, 0.0}
\definecolor{teal}{rgb}{0.0, 0.4, 0.4}
\definecolor{purple}{rgb}{0.65,0,0.65}
\definecolor{saffron}{rgb}{0.95,0.75,0.2}
\definecolor{turquoise}{rgb}{0.0,0.5,0.5}

\usepackage[ruled,vlined,linesnumbered]{algorithm2e}
\usepackage{multirow}

\usepackage[normalem]{ulem}

\usepackage{mathtools}
\usepackage{amsmath, amsthm, amssymb, amsfonts, amsopn}
\usepackage{graphicx} 
\usepackage{textcomp}
\usepackage{xfrac}
\usepackage{bbm}

\usepackage{overpic}
\usepackage{subfig}
\usepackage{wrapfig}

\newcommand{\hidecomment}[1]{}

\renewcommand{\vec}[1]{\mathbf{#1}}

\newcommand{\etal}[1]{et al.}
\renewcommand{\raggedright}{\leftskip=0pt \rightskip=0pt plus 0cm}

\begin{document}
%
\title{Edge Preserving Implicit Surface Representation of Point Clouds} 

\author{Xiaogang Wang, Yuhang Cheng, Liang Wang, Jiangbo Lu,~\IEEEmembership{Senior Member,~IEEE}\;,  Kai Xu,~\IEEEmembership{Senior Member,~IEEE}\;, Guoqiang Xiao
\IEEEcompsocitemizethanks{\IEEEcompsocthanksitem Xiaogang Wang, Yuhang Cheng, Liang Wang and Guoqiang Xiao are with College of Computer and Information Science, Southwest University, China. 
\IEEEcompsocthanksitem Jiangbo Lu is with the SmartMore Co., Ltd.
\IEEEcompsocthanksitem Kai Xu is with the National University of Defense Technology, China.
}
}

\markboth{Journal of \LaTeX\ Class Files, ~2019}%
{Shell \MakeLowercase{\textit{et al.}}: Bare Demo of IEEEtran.cls for Computer Society Journals}
%



\IEEEtitleabstractindextext{%
\begin{abstract}
\raggedright{
Learning implicit surface directly from raw data recently has become a very  attractive representation method for 3D reconstruction tasks due to its excellent performance. However, as the raw data quality deteriorates, the implicit functions often lead to unsatisfactory reconstruction results.
To this end, we propose a novel edge-preserving implicit surface reconstruction method, which mainly consists of a differentiable Laplican regularizer and a dynamic edge sampling strategy. Among them, the differential Laplican regularizer can effectively alleviate the implicit surface unsmoothness caused by the point cloud quality deteriorates; Meanwhile, in order to reduce the excessive smoothing at the edge regions of implicit suface, we proposed a dynamic edge extract strategy for sampling near the sharp edge of point cloud, which can effectively avoid the Laplacian regularizer from smoothing all regions. Finally, we combine them with a simple regularization term for robust implicit surface reconstruction. Compared with the state-of-the-art methods, experimental results show that our method significantly improves the quality of 3D reconstruction results.
Moreover, we demonstrate through several experiments that our method can be conveniently and effectively applied to some point cloud analysis tasks, including point cloud edge feature extraction, normal estimation,etc.%

}
\end{abstract}

}

\maketitle

\IEEEdisplaynontitleabstractindextext

%
\IEEEpeerreviewmaketitle

\pdfoutput=1
\pdfoutput=1
\pdfoutput=1
\pdfoutput=1
\pdfoutput=1
\section{Introduction}
\pdfoutput=1

Recently, Implicit Neural Representations (INRs) has gained made great strides in the field of 3D reconstruction 
~\cite{chen2019learning,mescheder2019occupancy,park2019deepsdf,chibane2020implicit, erler2020points2surf,peng2020convolutional,saito2019pifu,xu2019disn}. 
In contrast to traditional explicit representations such as point clouds ~\cite{fan2017point}, voxels~\cite{choy20163d, wu2016learning} and mesh ~\cite{groueix1802atlasnet,kato2018neural,tang2019skeleton,tang2021skeletonnet}, implicit neural representations represent surface function primarily through neural networks, providing higher quality, flexibility, and fidelity without discretization errors, and significantly save amounts of storage space to store high-quality results.

However, most of these methods need ground truth data as supervision~\cite{chen2019learning, mescheder2019occupancy,park2019deepsdf}, which have difficulty in generalizing well to unseen shapes that are dissimilar to the training samples.
Recently, some methods ~\cite{atzmon2019controlling,atzmon2020sald,atzmon2020sal,gropp2020implicit,zhao2021sign} have been proposed to reconstruct implicit neural representations directly from raw data (point clouds, triangle soups, unoriented meshes, etc.).
Compared to data-driven approaches, building implicit neural representations directly from raw data is obviously more appealing.
Generally speaking, the core idea of such methods is to impose explicit/implicit regularity constraints to reduce reliance on dataset. 
SAL ~\cite{atzmon2020sal} proposed a unsigned regression loss to a given unsigned distance function to raw data, which can produce signed solutions of implicit functions.
\begin{figure}[h!]
\begin{center}
\includegraphics[width=0.46\textwidth]{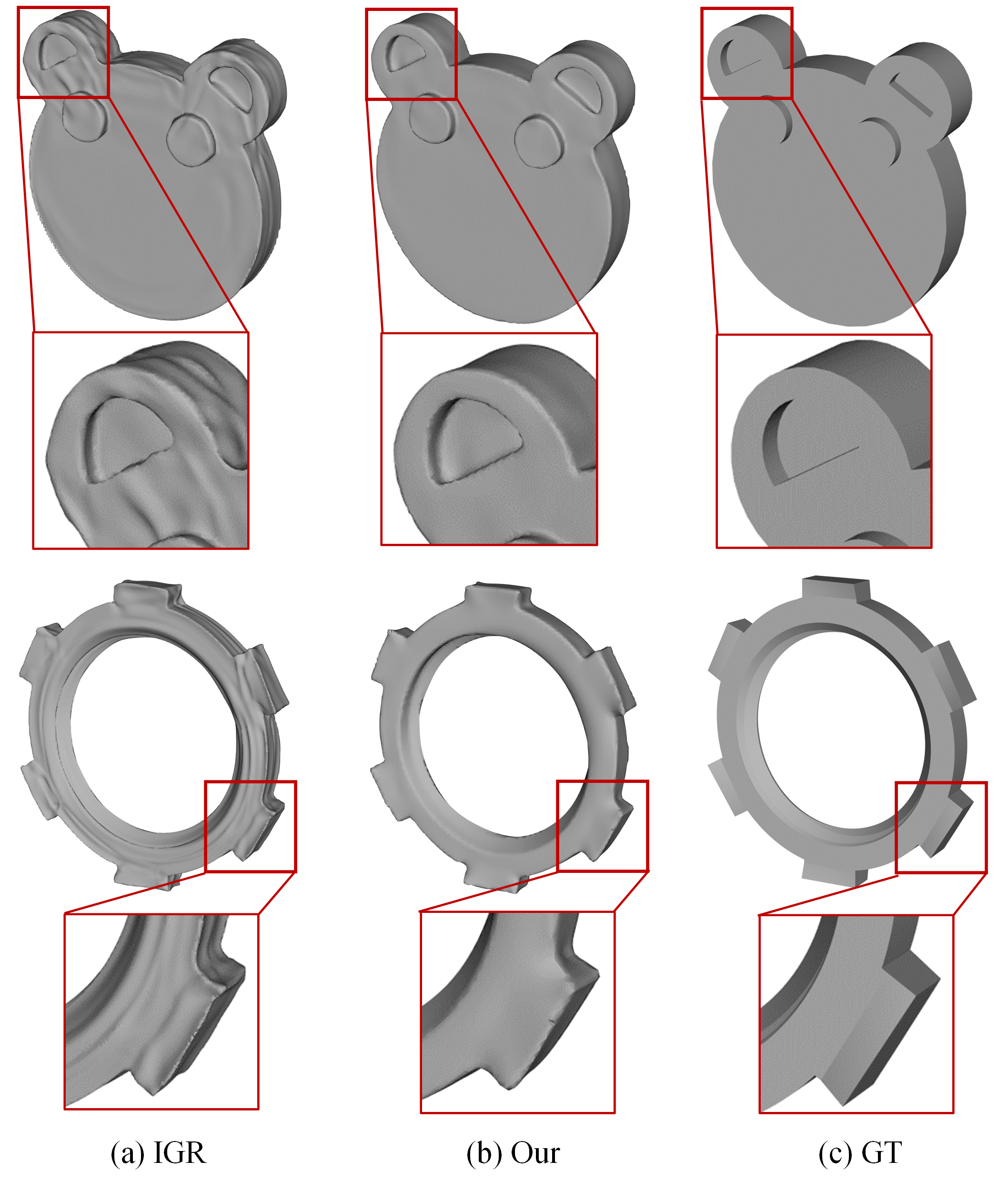}
\end{center}
\caption{Effect of edge preserving differential Laplacian regularizer. (b) are the optimization results of the edge preserving differential Laplacian regularizer which is incorporated into the state-of-the-art methods IGR~\cite{gropp2020implicit} . (a) and (c) are the results of IGR and Ground truth respectively.}
\label{fig:teaser}\vspace{-8pt}
\end{figure}
Specifically, starting from raw data (e.g., point clouds, real scanned grids, etc.), implicit neural representations learn in a self-supervised manner and can be trained reliably relying only on raw input data by minimizing unsigned regression.
Subsequently, SALD ~\cite{atzmon2020sald}, a generalized version of SAL ~\cite{atzmon2020sal} was proposed, which can obtain higher quality reconstruction results by incorporating an explicit gradient constraint on SAL.
Gropp et al. ~\cite{gropp2020implicit} proposed a novel implicit geometric regularization (IGR) method  to directly learn an implicit neural representation from raw data and achieved surprising results.
Different from SAL~\cite{atzmon2020sal} and SALD~\cite{atzmon2020sald}, IGR only relies on implicit regularization constraints, without the need for a unsigned distance function.
More specifically, IGR proposes an implicit geometric regularization, which amounts to solving a particular Eikonal boundary value problem that constrains the norm of spatial gradients to be $1$ almost everywhere.
Yet, when the normal information cannot be available and the number of input points is not dense enough, the above algorithms often lead to unsatisfactory reconstruction results (See Figure~\ref{fig:teaser}(a)).

We observed that the main reason for the unsatisfactory reconstruction results is that the implicit function needs to fit the input point cloud as much as possible, and the noise information in the point cloud tends to cause the implicit surface to be very unsmooth.
In other words, the main reason for this phenomenon is the inconsistency of normal in the local region of the reconstructed surface. Therefore, it is an intuitive idea to keep the local normal of the surface consistent as much as possible; Meanwhile, it should be noted that not all regions are restricted in their normal consistency, for example, obviously sharp edges often exist in the surface (as shown in Figure~\ref{fig:teaser}). In the reconstruction process, we hope that this part of the area will not be overly smoothed. Therefore, The edge preserving local normal consistency is more accurate for implicit surface representation .

In view of the above problem, it can be visually viewed as a standard Laplacian minimization problem; Meanwhile, we can also use the Laplacian operator to identify the edge region effectively, which has achieved good results in many image processing tasks. Therefore, in other words, we can design an intuitive Laplacian regularization, which can effectively improve the quality of reconstruction results.

However, in this task, the raw data type we consider is point cloud data, and the difference method cannot be directly used to approximate high-order derivatives, mainly because point cloud data does not have a clear topological relationship like mesh or image.
If the algorithm similar to KNN is used, the nearest neighbor points searched cannot guarantee the correct topology structure
(as shown in Figure~\ref{fig:diff_top}), especially when the point cloud is not dense and the normal are not available 
, such wrong nearest neighbor results will easily lead to the anti-optimization results (as shown in Figure~\ref{fig:over_L_neibor_L_laplace}(a)).

Recently, there is growing interest in differentiable optimization of implicit neural representations that enable differential nature as supervision in learning frameworks 
~\cite{park2019deepsdf,sitzmann2020implicit,gropp2020implicit,chen2022implicit,
yariv2021volume,ehret2022nerf,zehnder2021ntopo}.
The advantage of differentiable implicit neural representations is that it can directly solve the higher derivative of the input signal instead of discretization approximation, which greatly improves its optimization performance and application range.
Thanks to the analytically-differentiable nature of implicit neural representation, we can easily design a differentiable Laplacian regularizer.
Meanwhile, the differentiable Laplacian regularizer can be easily and intuitively incorporated into implicit neural surface representations (as shown in Figure~\ref{fig:teaser}). We show that it significantly improve the quality of 3D reconstruction.
Meanwhile, in order to facilitate qualitative and quantitative comparisons in this paper, unless otherwise stated,
in this paper, all experimental results are obtained by incorporating them into IGR ~\cite{gropp2020implicit}. We carefully evaluate its performance through a series of ablation studies. Meanwhile, we demonstrate through several experiments that our method can be conveniently and effectively applied to some point cloud analysis tasks, including point cloud edge feature extraction, normal estimation, etc.

In summary, we make the following contributions: In this paper, we use the infinite differentiability property of implicit neural representation to propose a novel edge-preserving implicit surface reconstruction method, which mainly consists of a differentiable Laplican regularizer and a dynamic edge sampling strategy. 1), Among them, the differential Laplican regularizer can effectively alleviate the implicit surface unsmoothness caused by the point cloud quality deteriorates; 2), Meanwhile, in order to reduce the excessive smoothing at the edge regions of implicit suface, we proposed a dynamic edge extract strategy for sampling near the sharp edge of point cloud, which can effectively avoid the Laplacian regularizer from smoothing all regions.

\pdfoutput=1
\pdfoutput=1
\pdfoutput=1
\pdfoutput=1
\pdfoutput=1
\section{Related Work}

\subsection{Data-driven based Implicit surface reconstruction}
\label{subsec:Data-driven based Implicit surface representation}

3D surface reconstruction from raw data has gained significant research progress in recent year, benefiting from the advances in machine learning techniques ~\cite{chen2019learning,mescheder2019occupancy,park2019deepsdf,chibane2020implicit, erler2020points2surf,peng2020convolutional,saito2019pifu,xu2019disn}. 
Early studies ~\cite{carr2001reconstruction,alexa2003computing,kazhdan2006poisson} most utilize predefined geometric priors (such as local linearity and smoothness) towards specific tasks. These geometric priors often encode statistical properties of raw data and are designed to be optimized, such as poisson equation ~\cite{kazhdan2006poisson,kazhdan2013screened}, radius basis function ~\cite{carr2001reconstruction}, moving least squares ~\cite{alexa2003computing}.
Recently, 
implicit neural representation has gained significant research progress for geometry reconstruction
~\cite{genova2019learning,chen2019learning,atzmon2019controlling,mescheder2019occupancy,
niemeyer2019occupancy,oechsle2021unisurf,park2019deepsdf,peng2020convolutional,saito2019pifu,tancik2020fourier,
wang2021neus} and object representation ~\cite{park2019deepsdf, sitzmann2019scene, yariv2021volume,bergman2021fast, chibane2021stereo, gao2020portrait, jiang2020sdfdiff, kellnhofer2021neural, liu2020neural, martin2021nerf, mildenhall2020nerf,niemeyer2020differentiable,yariv2020multiview} 
due to their simplicity and excellent performance, which learn an approximate implicit function with multi-layer perceptron (MLP).
Compared to the traditional continuous and discrete representations (grid, point cloud and voxel), implicit neural representations have many potential benefits, which can provide higher modeling quality without discretization errors, flexibility and fidelity, and save storage space.
However, most of these methods need ground truth data as supervision  
~\cite{chen2019learning,mescheder2019occupancy,park2019deepsdf}, which have difficulty in generalizing well to unseen shapes that are dissimilar to the training samples.

In addition, there are hybridization-based methods ~\cite{jiang2020local,tretschk2020patchnets, chabra2020deep,yang2021deep,tang2021sa} that combine data-driven priors with optimization strategy that can achieve state-of-the-art performance. However, the above methods also require additional ground truth data as supervision, which seriously limits their applicability. 

\subsection{Sign Agnostic Implicit surface reconstruction}
\label{subsec:Sign Agnostic Implicit surface reconstruction}
 
Recently, some methods~\cite{atzmon2020sald,atzmon2020sal,gropp2020implicit,zhao2021sign} have been proposed to reconstruct implicit neural representations directly from raw data. Compared to big data-driven approaches, building implicit neural representations directly from raw data is obviously more appealing. These methods can avoid the need for a large number of ground truth signed distance representation of  training data as supervision. SAL ~\cite{atzmon2020sal} introduces a sign agnostic regression loss to a given unsigned distance function to raw data, which is the signed version of unsigned distance function. Meanwhile, that avoids the use of surface normals by properly initializing implicit decoder networks so that they can only produce signed solutions of implicit functions using unsigned distance function. Subsequently, SALD ~\cite{atzmon2020sald}, a generalized version of SAL ~\cite{atzmon2020sal} was proposed, which can obtain higher quality reconstruction results by incorporating an explicit gradient constraint on SAL. 
Similarly, in this paper, our approach also uses implicit neural representation to estimate level set functions directly from raw data. The major difference is that our proposed regularization terms are directly based on differentiable implicit optimization, and does not explicitly enforce some regularization on the zero level set, such constraints, when the normal information cannot be available and the number of input point cloud is not dense enough, the implicit neural representation often lead to unsatisfactory reconstruction results.

\subsection{Differentiable implicit neural representation}
\label{subsec:Differentiable Implicit neural representation}

Compared with general implicit neural representation, differentiable implicit neural representation has the advantage that it can directly use various properties of differential geometry instead of discretization approximation, which can lead to more stable solutions in many optimization problems.
Recently, there is growing interest in differentiable optimization of implicit neural representation that enable differential nature as supervision in learning frameworks 
~\cite{park2019deepsdf,sitzmann2020implicit,gropp2020implicit,sitzmann2020implicit,chen2022implicit,
yariv2021volume,ehret2022nerf,zehnder2021ntopo}.
General numerical optimization often uses the discrete approximation of differential geometry, 
for example, finite difference method is often used to enhance the smoothness between adjacent samples in space.
But thanks to the analytically-differentiable nature of implicit neural representation, 
differentiable implicit neural representations can make direct use of many properties in differential geometry, such as gradients ~\cite{gropp2020implicit,sitzmann2020implicit,yariv2021volume}, curvatures ~\cite{ehret2022nerf}, and the solution of partial differential equations ~\cite{chen2022implicit,zehnder2021ntopo}.
Recently, 
Gropp et al.~\cite{gropp2020implicit} proposed to use the differentiable implicit neural representation to directly reconstruct  surface from raw data. More specifically, it proposes an implicit regularization constraint, which amounts to solving a particular Eikonal boundary value problem that constrains the norm of spatial gradients to be $1$ almost everywhere. Similarly, Sitzmann et al. ~\cite{sitzmann2020implicit} uses the proposed a differentiable periodic activation functions to represent signed distance fields in a fully-differentiable manner.
Both of these works~\cite{gropp2020implicit,sitzmann2020implicit} , however, when the normal information cannot be available and the number of input points is not dense enough, often lead to unsatisfactory reconstruction results.
In this paper, our work is also based on the differentiability of implicit neural representations to optimize implicit level set function estimated directly from the input point cloud.
Specifically, we designed an implicit differentiable Laplacian regularizer, which effectively alleviated the problem of unsatisfactory reconstruction results caused by direct fitting of input point cloud by implicit neural function.


\pdfoutput=1
\pdfoutput=1
\pdfoutput=1
\pdfoutput=1
\pdfoutput=1
\section{Method}
\label{sec:method}

We present a differentiable laplacian regularizer for neural implicit representation directly from input point cloud without normal supervision. 
Note that our differential Laplacian regularizer can be incorporated into any implicit neural representation, such as IGR~\cite{gropp2020implicit},SAL~\cite{atzmon2020sal},SALD~\cite{atzmon2020sald}.
In this paper, unless otherwise noted, we incorporate it in the IGR, which use level sets of neural network to represent 3D shape (Sec. 3.1). 
More specifically, IGR proposes an implicit geometric regularization, which amounts to solving a particular Eikonal
boundary value problem that constrains the norm of spatial gradients to be 1 almost everywhere. Yet, when the normal information cannot be available and the number of input points is not dense enough, IGR often lead to unsatisfactory reconstruction results (See Figure~\ref{fig:teaser}(a)).
We observed that the main reason for the unsatisfactory reconstruction results is that the implicit function needs to fit the input point cloud as much as possible, and the noise information in the point cloud tends to cause the implicit surface to be very unsmooth.

To overcome this problem, we use the analytically-differentiable nature of implicit neural representation,
to propose a differential Laplacian regularizer, which can effectively alleviate the unsatisfactory reconstruction results (Sec. 3.2).
Meanwhile, in order to reduce the excessive smoothing at the edge regions of 3D shape (such as man-made shapes), a dynamic
edge extraction strategy (Sec. 3.2) is introduced for sampling near the sharp edge of input point cloud, which can effectively avoid the Laplacian regularizer from smoothing all regions, so as to effectively improve the quality of reconstruction results while maintaining the edge.

\subsection{Background}
\label{subsec:Background}
A neural implicit representations is a continuous function that approximate the signed distance function. 
The underlying surface of 3D shape is implicitly represented by the zero level set of this function, 
\begin{equation}
  f_{\theta}(x) = 0, \forall x\in X.
\label{eq:fun_1}
\end{equation}
where $\theta$ indicates the parameters to be learned and $X$ indicates the set of input point cloud.
In general, one parameterize this function using a multi-layer perceptron (MLP). Meanwhile, in order to conveniently use the analytically-differentiable (such as, gradients,etc.) nature of implicit neural representation, recent works ~\cite{gropp2020implicit,sitzmann2020implicit} usually replace the commonly used ReLU activation function with a non-linear differentiable activation functions, thus transforming MLP into a continuous and infinitely differentiable function.  

\begin{figure}[t]
  \centering
  \begin{overpic}[width=\linewidth,tics=10]{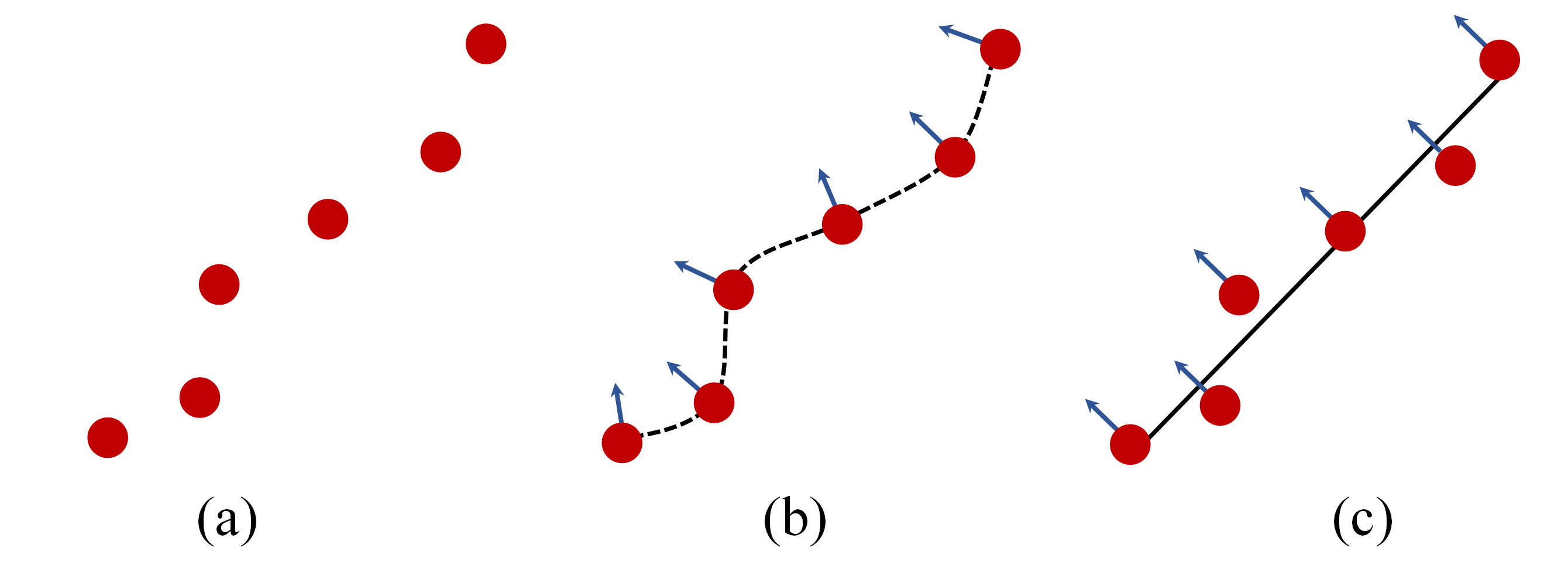}
  \end{overpic}
  \caption{Illustrations of the local normal consistency.}
  \label{fig:norm_cosist}
  \vspace{-8pt}
\end{figure}

In IGR, the training is done by minimizing the loss that encourages $f$ to vanish on $X$:
\begin{equation}
  L_{vanish} = \frac{1}{N(X)}\sum_{x \in X}{|f_\theta(x)|}
\label{eq:L_van}
\end{equation}
where $N(X)$ is the number of point set $X$, $|\bullet|$ indicates absolute value.
if the input point cloud includes normal information $\vec{n_{gt}(x)}$, the corresponding loss function can be designed to make the predicted normal (the differentiable gradient $\triangledown{f_\theta(x)}$ of the implicit function) as close as possible to the ground truth normal $\vec{n_{gt}(x)}$:
\begin{equation}
  L_{normal} = \frac{1}{N(X)}\sum_{x \in X}{||\triangledown{f_\theta(x)}-\vec{n_{gt}(x)}||_2}
\label{eq:L_normal}
\end{equation}

In addition to the above two intuitive fitting loss terms, IGR ~\cite{gropp2020implicit} based on the Eikonal partial differential equation presents an additional loss (Eikonal loss), which is equivalent to solve boundary value problems of a particular Eikonal that constrains the norm of spatial gradients $\triangledown{f_\theta(x)}$ to be $1$ almost everywhere:
\begin{equation}
  L_{eikonal} = \frac{1}{N(X)}\sum_{x \in X}{(||\triangledown{f_\theta(x)}||_2-1)^2}
\label{eq:L_eikonal}
\end{equation}

Note that, in our approach, we do not consider normal information as supervision, so we will not consider $L_{normal}$ term in all subsequent experiments. More specifically, our approach builds upon the above two items $L_{vanish}$ and $L_{eikonal}$.

\begin{figure}[t]
  \centering
  \begin{overpic}[width=\linewidth,tics=10]{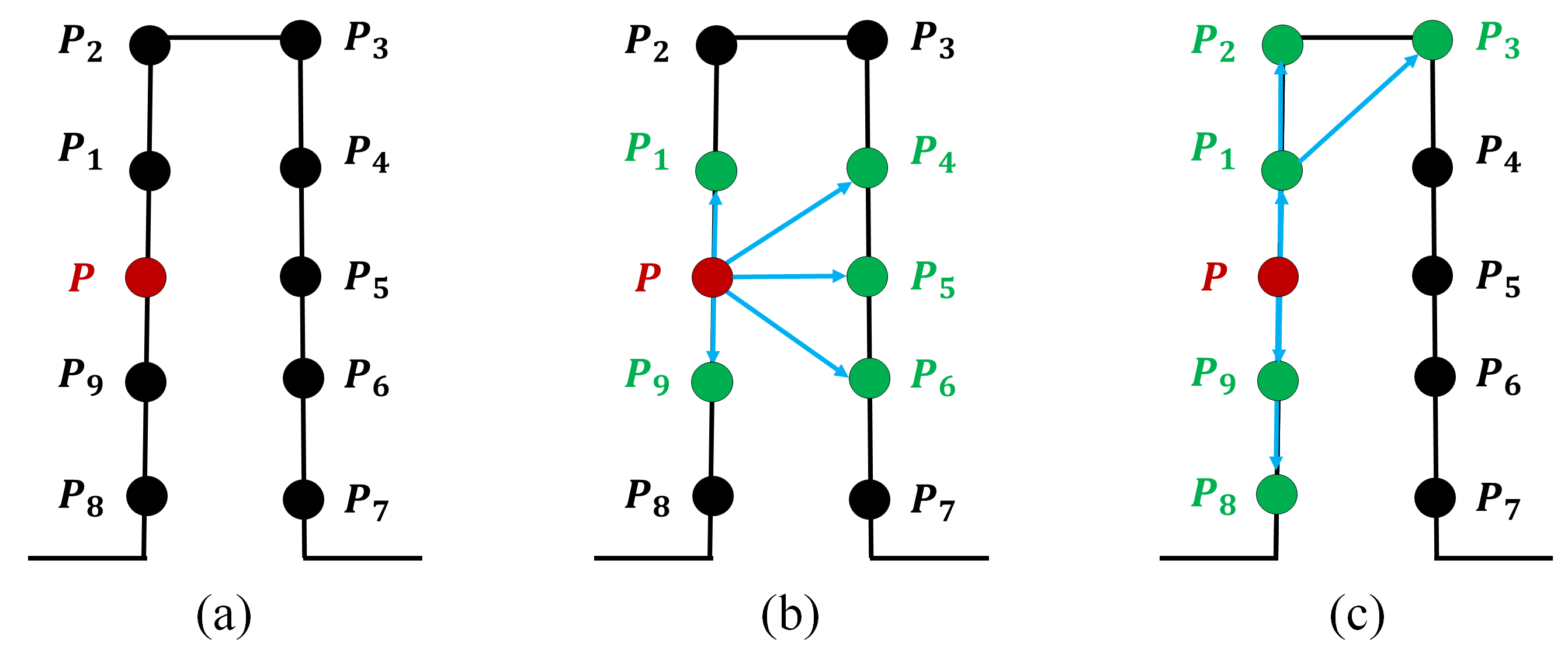}
  \end{overpic}
  \caption{Illustrations of two different $N$ nearest neighbors of non-topological preservation (b) and topological preservation (c) for geometric structure (a).}
  \label{fig:diff_top}
  \vspace{-8pt}
\end{figure}

\subsection{Differentiable laplace regularization}
\label{subsec:Differentiable laplace regularization}

\textbf{Neighborhood normal consistency.}
A high-quality result can be generated based on the above two terms ($L_{vanish} $ and $L_{eikonal}$) when the input point data is large enough, however, when the normal information cannot be available and the number of input points is not dense enough, often lead to unsatisfactory reconstruction results (See Figure~\ref{fig:teaser}(a)).

We observed that the main reason for the unsatisfactory reconstruction results is that the implicit function needs to fit the input point cloud as much as possible, and the noise information in the point cloud tends to cause the implicit surface to be very unsmooth. 
More specifically, the optimization results are not guaranteed to provide a high-quality reconstruction result, which is intuitively reflected by the possibility that the normal of reconstruction result is inconsistent in the neighborhood.


From another perspective, it is well known that 3D shapes tend to be piecewise smooth, that is, flat surfaces are more likely than high-frequency structures ~\cite{huang2000statistics}.  For this purpose, we incorporate this prior into implicit neural function by encouraging the geometric smoothness of the reconstructed results. 
Therefore, an intuitive solution is to constrain the consistency of the neighborhood normal of the reconstruction results (as shown in Figure~\ref{fig:norm_cosist}):

\begin{equation}
  L_{neibor} = \sum_{x \in X}\sum_{x_i \in nei(x)} {||\triangledown{f_\theta(x)} - \triangledown{f_\theta(x_i)}||_2}
\label{eq:L_neibor}
\end{equation}
where $nei(x)$ indicates the neighbor point set of point $x$.

\begin{figure}[t]
  \centering
  \includegraphics[width=1\linewidth]{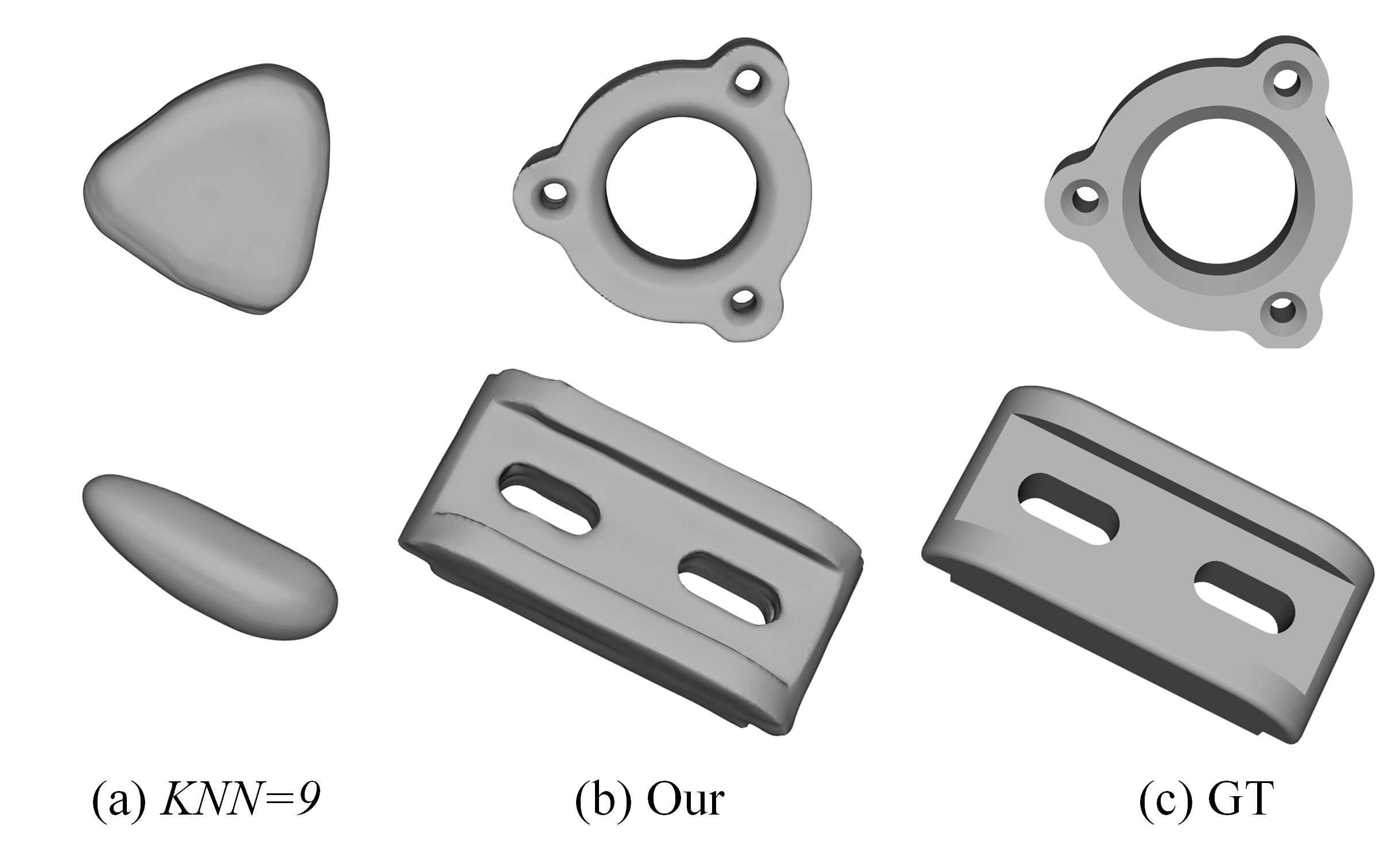}
  \caption{The comparison of $L_{neibor}$ (a) and  $L_{laplacian}$ (b).}
  \label{fig:over_L_neibor_L_laplace}
  \vspace{-8pt}
\end{figure}

However, in this paper, the raw data type we consider is point cloud data, which does not have a clear topological structure like mesh or voxels. 
If the algorithm similar to KNN is used, the nearest neighbor points searched cannot guarantee that they maintain the correct topology structure, especially when the point cloud is not dense and the normal are not available,  
as shown in Figure~\ref{fig:diff_top}(b) where the three points $P_4$, $P_5$ and $P_6$ do not meet the nearest neighbor result of $N=5$ under the maintenance of the topology structure, and the correct set of nearest neighbor points should be  $\{P_1,P_2,P_3,P_8,P_9\}$. Moreover, it is difficult to get a reasonable value for this parameter $nei(x)$ in practice.
As shown in Figure~\ref{fig:over_L_neibor_L_laplace}, we can easily see that the wrong reconstructed results, which is mainly caused by the above reasons. 
 


\textbf{Differentiable Laplacian regularizer.}
In fact, the above constraint $L_{neibor}$ is mainly used to constrain the normal consistency in the local domain, which can be easily interpreted as a discrete Laplace operator.
The Laplacian operator $\triangle f$ is a second-order differential operator in $n$-dimensional euclidean space, defined as the divergence ($\triangledown \cdot f$) of the gradient ($\triangledown{f}$).
Thanks to the infinite differentiability of implicit neural representation, we can design a simple but effective differentiable Laplacian regularizer:

\begin{equation}
  L_{laplacian} = \sum_{x \in X}{\triangle{f_\theta(x)}^2}
\label{eq:L_laplace}
\end{equation}
where $\triangle{f_\theta(x)}$ indicates the differentiable Laplace operator of point $x$.

As shown in Figure~\ref{fig:over_L_neibor_L_laplace}(b), compared with the explicit regularization constraint $L_{neibor}$ based on the nearest neighbor normal consistency, the differentiable Laplacian regularizer can obtain more stable results without introducing hyperparameter nearest neighbors $N$.

\subsection{Dynamic edge sampling}
\label{subsec:Dynamic edge sampling}
\begin{figure}[t]
  \centering
  \includegraphics[width=1\linewidth]{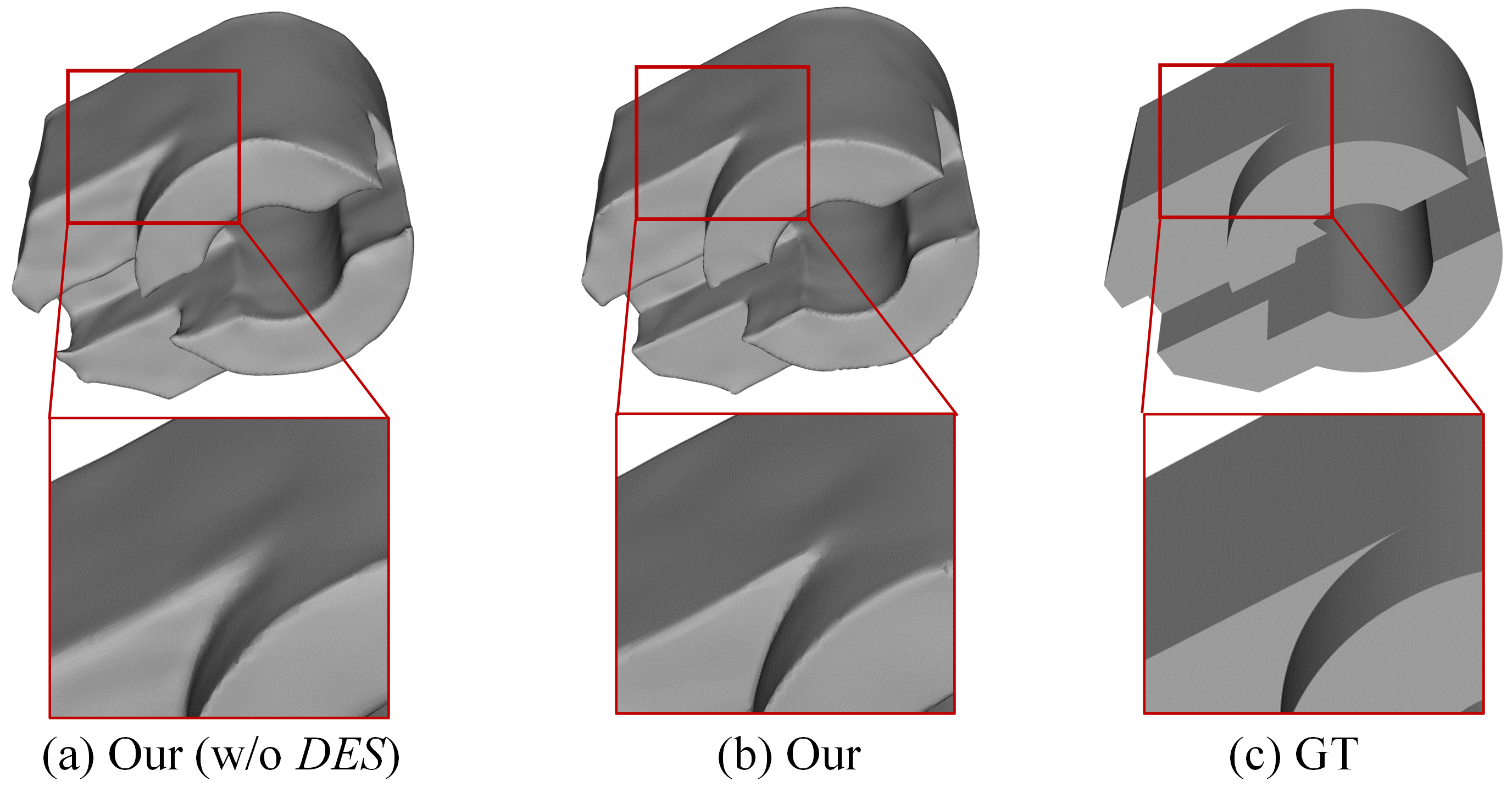}
  \caption{The comparison of with (b) and without $Dynamic$ $Edge$ $Sampling$ $(DES)$ (a).}
  \label{fig:over_dynamic_edge_sampling}
  \vspace{-8pt}
\end{figure}

However, while the differentiable Laplacian regularizer restricts the normal consistency, it also brings a new problem: 
It imposes undifferentiated constraints on all 3D regions, even in the sharp-edge regions, 
as shown in Figure~\ref{fig:over_dynamic_edge_sampling}.
As we know, complex 3D shapes are generally constructed by multiple piecewise smooth surfaces, which may not be differentiable at the joints, and are more likely to form sharp edges.
Therefore, in essence, a complex 3D shape (piecewise smooth model with sharp edges) cannot be accurately represented by an implicit function, because it is obviously not differentiable at sharp edges, so if it is forced to be represented by an implicit function, especially only sparse point sets without  normal information are used as supervision, it is easy to form an overly smooth reconstruction at the sharp edges (as shown in Figure~\ref{fig:over_dynamic_edge_sampling}).

The most intuitive solution is to implicitly represent each piecewise smooth surface separately, but this is difficult to do in practice because it first requires the segmentation of the input point set, which is difficult to do accurately in unsupervised conditions.
 
Therefore, we propose a novel $dynamic$ $edge$ $sampling$ strategy to effectively extract sharp edge regions in the training process. In theory, the remaining regions not only satisfy the differentiable property, but also conform to the normal consistency constraint, which can effectively avoid the indifference smoothing of all regions, including the edge regions, of the laplace regularizer.

\begin{figure}[t]
  \centering
  \includegraphics[width=1\linewidth]{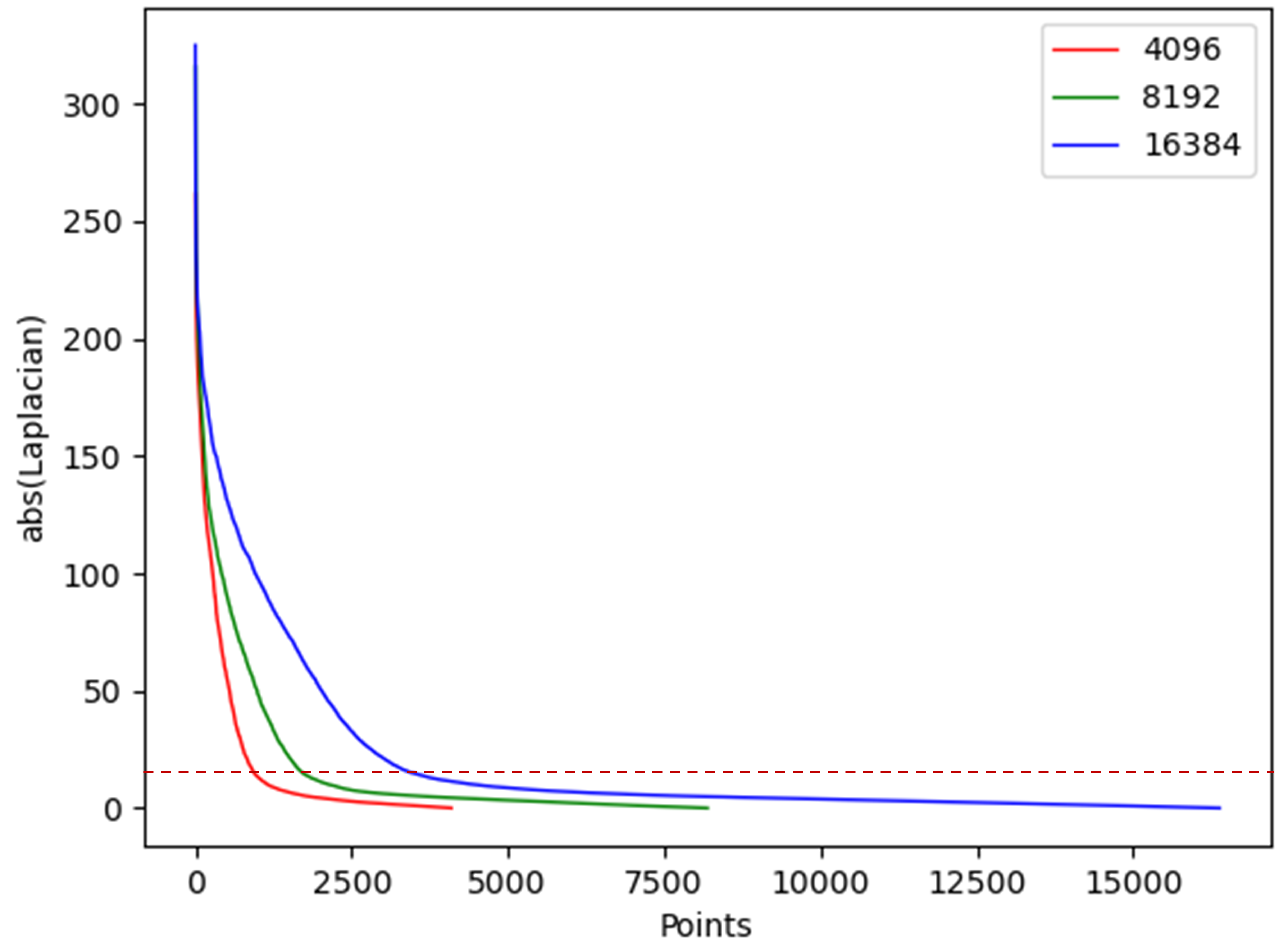}
  \caption{Statistics of Laplacian operators $|\triangle{f_\theta(x)}|$ and edge threshold $\tau$ selection.}
  \label{fig:edge_detection_thres}
  \vspace{-8pt}
\end{figure}

Specifically, for each point $p$ in the input point set, we may quickly determine whether it is an edge point according to its differentiable Laplacian operator $\triangle{f_\theta(x)}$. Essentially, Laplacian  is mainly used to describe the rate of change of gradient, and is often used for edge detection in image processing. From the perspective of differential geometry, it is used to describe the change rate of spatial position normal. Therefore, the larger the laplacian of the point, the stronger the possibility that the point is an edge point.
We threshold the Laplacian  $|\triangle{f_\theta(x)}|<\tau$  to obtain a corresponding set of non-edge points $X'$.
According to statistics (as shown in Figure~\ref{fig:edge_detection_thres}), we set the parameter $\tau = 20$ throughout our experiments. 
This operation is performed before the backpropagation of each iteration, therefore, we call it $dynamic$ $edge$ $sampling$.
\begin{equation}
  L_{laplacian} = \sum_{x \in X'}{\triangle{f_\theta(x)}^2}
\label{eq:L_laplace}
\end{equation}
where $X'$ indicates the non-edge subset of the input point cloud $X$. Finally, we optimize the total loss:
\begin{equation}
  L_{total} = L_{vanish} + \lambda_1L_{eikonal}+ \lambda_2L_{laplacian}
\label{eq:L_laplace+des}
\end{equation} 
In which, we set $\lambda_1 = 0.1$ and $\lambda_2 = 0.001$ throughout our experiments.

\pdfoutput=1
\pdfoutput=1
\pdfoutput=1
\pdfoutput=1
\pdfoutput=1
\section{Details, Results and Evaluations}
\label{sec:opti}

\begin{figure*}[t!]
  \centering
  \includegraphics[width=1.0\textwidth]{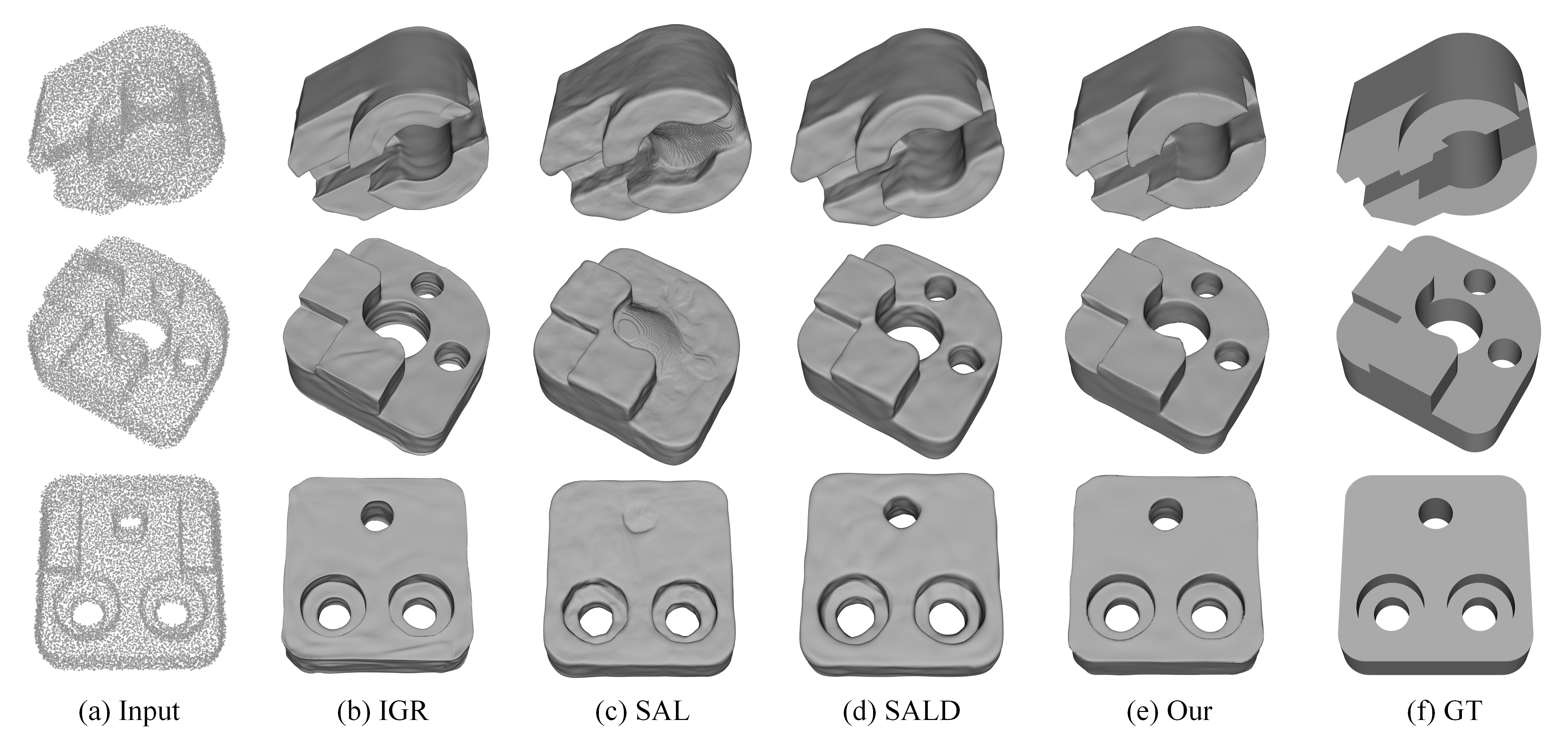}
  \caption{Qualitative comparison with state-of-the-art methods IGR~\cite{gropp2020implicit}, 
  SAL~\cite{atzmon2020sal} and SALD~\cite{atzmon2020sald}.}
  \label{fig:MoreResults_1}
  \vspace{-5pt}
\end{figure*}

\subsection{Implementation details}
\label{subsec:Implementation details}

\begin{figure}[t]
  \centering
  \includegraphics[width=1\linewidth]{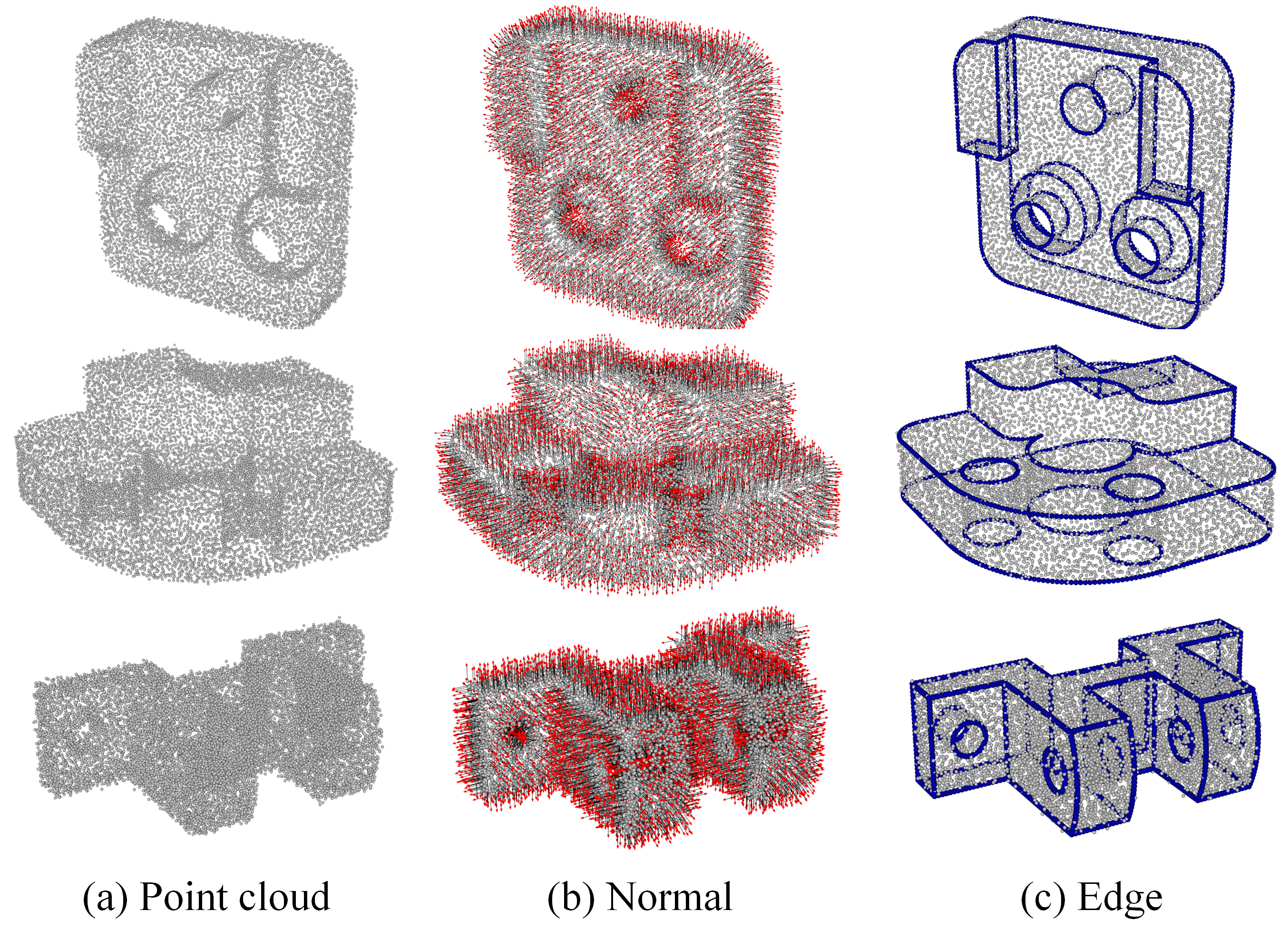}
  \caption{An overview of multi-task evaluation dataset.}   
  \label{fig:over_view_all}
  \vspace{-8pt}
\end{figure}

\begin{figure*}[ht]
  \centering
  \includegraphics[width=\textwidth]{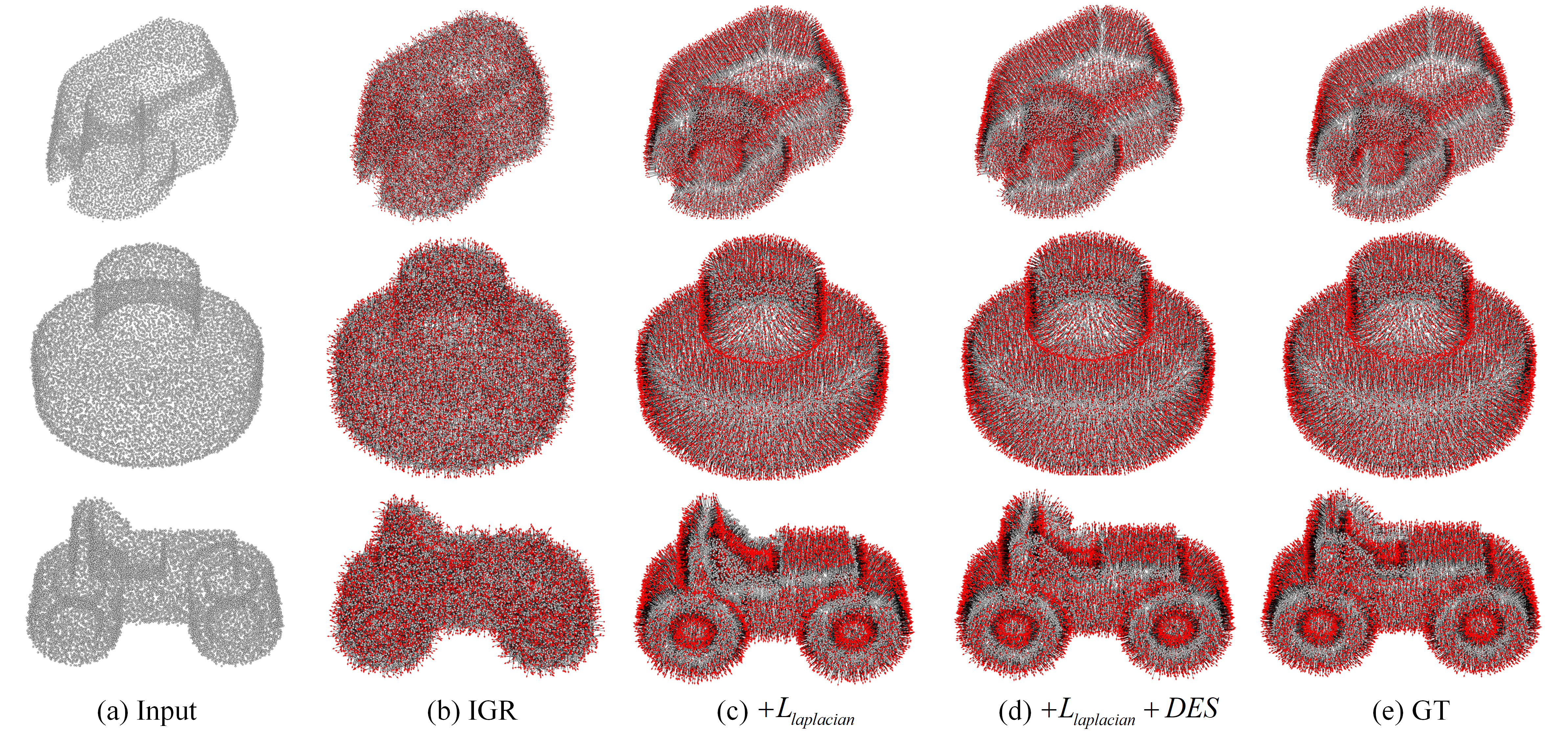}
  \caption{Visualization normal estimation of differential Laplacian regularizer (c) 
  and $dynamic$ $edge$ $sampling$ strategy (d).}
  \label{fig:Overview_optimization_net}
  \vspace{-8pt}
\end{figure*}

\begin{figure*}[ht]
  \centering
  \includegraphics[width=\textwidth]{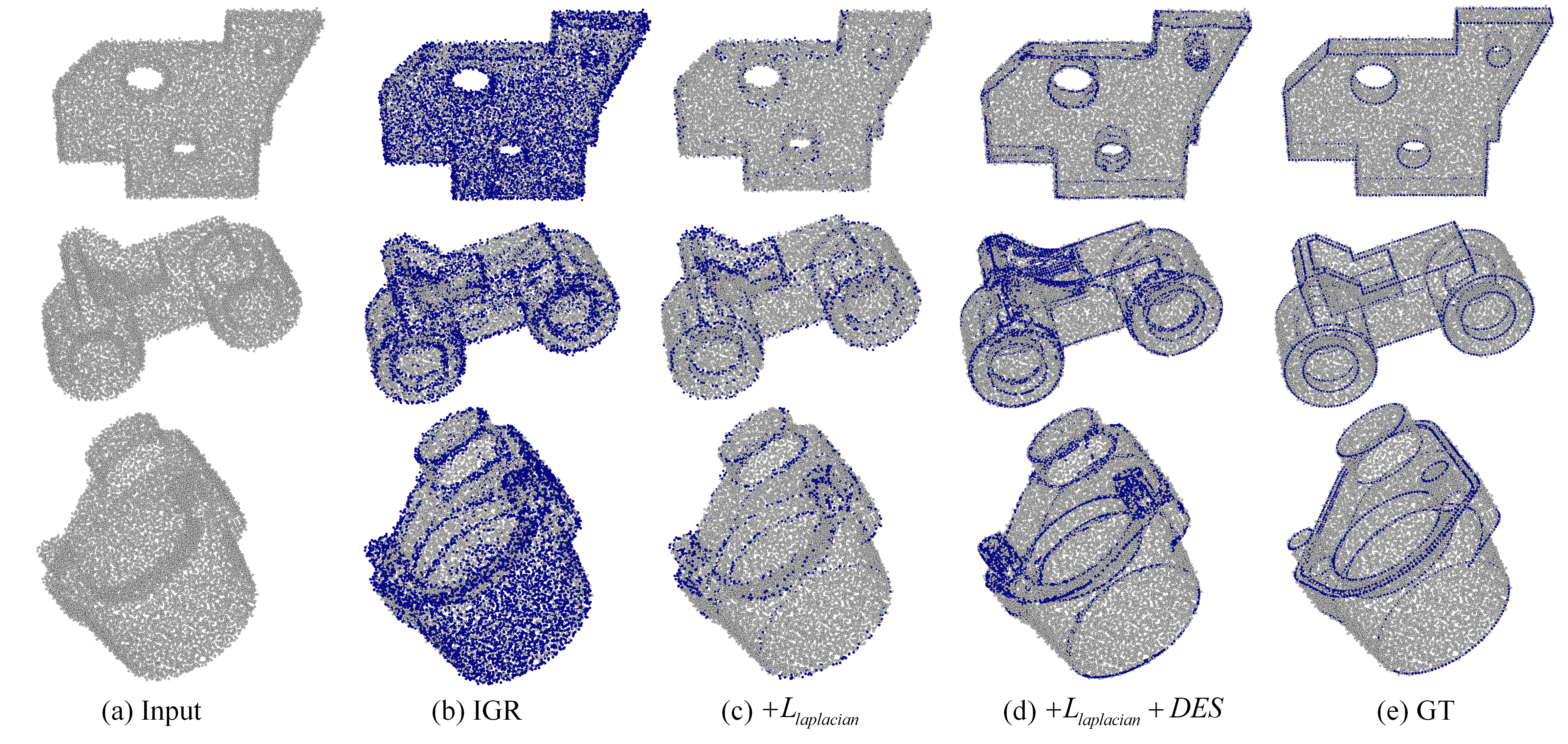}
  \caption{Visualization edge recognition of differential Laplacian regularizer (c) 
  and $dynamic$ $edge$ $sampling$ strategy (d).}
  \label{fig:edge}
\end{figure*}

\begin{figure}[t]
  \centering
  \includegraphics[width=1\linewidth]{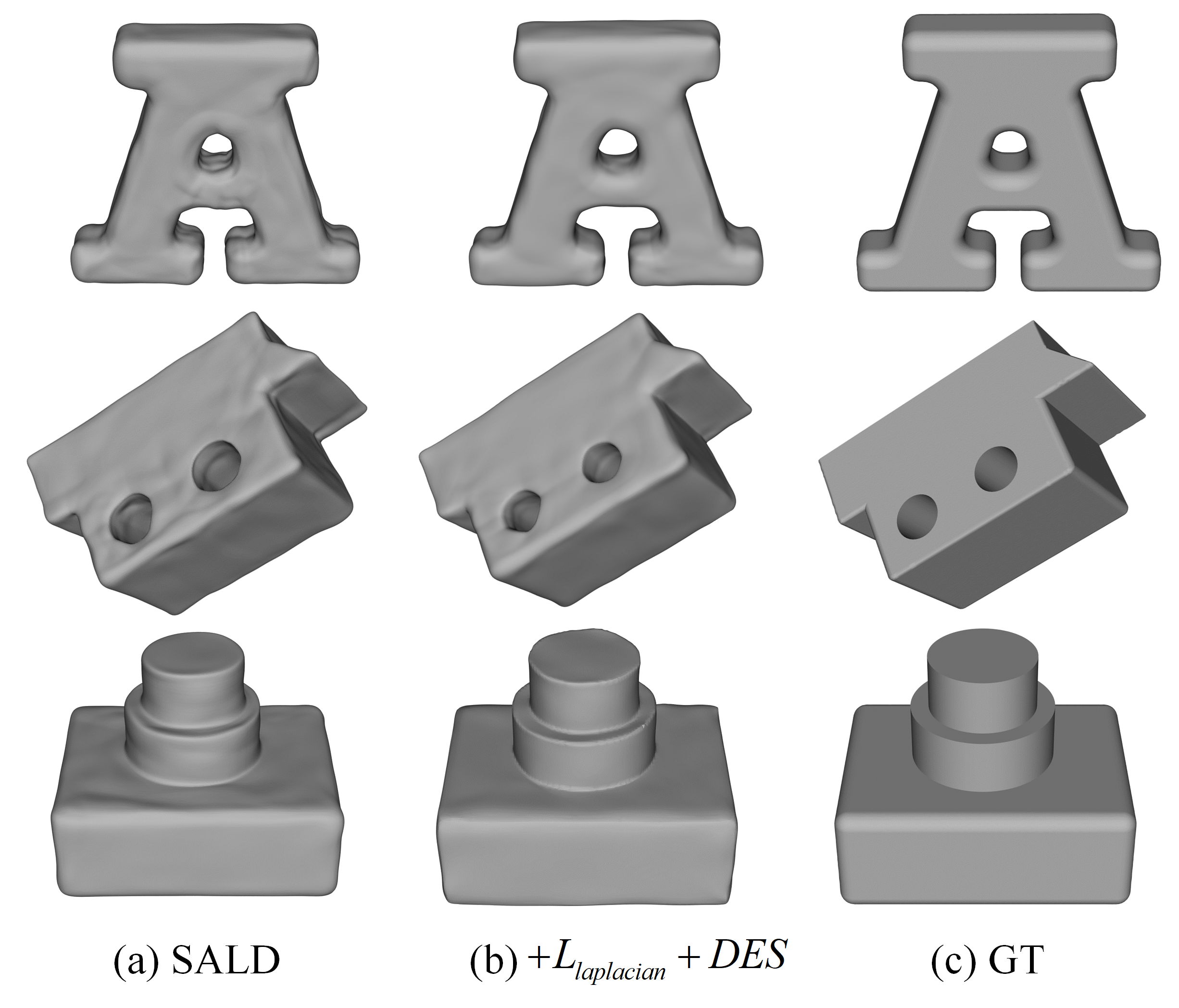}
  \caption{Effect of edge preserving differential Laplacian regularizer. (b) are the optimization results of the edge preserving differential Laplacian regularizer which is incorporated into the state-of-the-art method SALD~\cite{atzmon2020sald}. (a) and (c) are the results of SALD and Ground truth respectively.}
  \label{fig:SALD+LAP+DES}
  \vspace{-8pt}
\end{figure}

\textbf{Data preparation.}
To facilitate quantitative evaluation of our method on multiple tasks, including reconstruction , edge extraction and normal estimation, we selected 100 3D shapes with rich geometric topologies to construct the evaluation dataset 
(See Figure~\ref{fig:over_view_all}) from ABC dataset ~\cite{Koch_2019_CVPR}, which provides more than $1$ million standard 3D CAD models with multiple types of standard CAD format files. 
In addition to 3D geometry and normal information, the geometric edges information mentioned above does not provide us explicitly. To this end, we have developed a tool that, for each 3D shape, can quickly and easily extract the geometric edge information from the multiple CAD files, thus fully meeting the needs of our method for multi-task quantitative evaluation. 

\textbf{Point sampling.}
For each model, we sample it into a point cloud containing $16,384$ points by uniform point sampling. Meanwhile, in order to simulate the real point cloud noise, we added Gaussian noise with mean $\mu = 0$ and standard deviation $\delta = 0.005$ to each sampling point. In each case, except where otherwise stated, the network is trained on the noisy data throughout our experiments.
A few metrics on point cloud multi-tasks accuracy are defined to support quantitative evaluation of our approach; see the following subsections for details. 



\subsection{Metrics}
\label{subsec:Metrics}
In our experiments, both qualitative and quantitative evaluations are provided.
We evaluate our approach via ablation studies (Section 4.6), comparisons to state-of-the-art methods for 
3D reconstruction (Section 4.3) , edge detection (Section 4.4) and normal estimation (Section 4.5). 
For the quantitative assessment of the 3D reconstruction results, we used the two-sided Chamfer $d_C$ and Hausdorff distances $d_H$ introduced by ~\cite{gropp2020implicit}.
For the evaluation of the normal estimation, we use the angle $d_{angle}$ between the predicted normal and the groudtruth normal as the metric.
To evaluate edge detection, we measure $precision$/$recall$ and the $IoU$ between predictions and ground truth, while
to evaluate the geometric accuracy of the reconstructed edges, we employ the Edge Chamfer Distance (ECD) introduced by ~\cite{chen2019learning}. 

\subsection{Reconstruction}
\label{subsec:Reconstruction}
\textbf{Comparison with IGR~\cite{gropp2020implicit}.}
To facilitate a fair comparison with IGR~\cite{gropp2020implicit}, our network architecture is consistent with IGR~\cite{gropp2020implicit}. In all experiments, we used the default training procedure specified in IGR to train our network, except that we did not use normal information in the training and set iterations to $10000$. 
We set the loss parameters (see equation~\eqref{eq:L_laplace+des}) $\lambda_2=0.1$ and $\lambda_3=0.001$ throughout our experiments.
Qualitative and quantitative experiments are reported in Table~\ref{tab:compare_with_all_methods} and 
Figure~\ref{fig:MoreResults_1}
we can also see that the performance of our method is significantly better.


\noindent
\textbf{Comparison with state-of-the-art methods SAL~\cite{atzmon2020sal} and SALD ~\cite{atzmon2020sald}.}
In addition to IGR~\cite{gropp2020implicit}, our method is also compared with SAL ~\cite{atzmon2020sal} and SALD ~\cite{atzmon2020sald}, two state-of-the-art sign agnostic learning based methods from raw data.
The results shown in Table~\ref{tab:compare_with_all_methods}(row 1 and 2) are inferior to those of our method.
As shown in Figure~\ref{fig:MoreResults_1}, the results demonstrate the significant advantage of our approach, due to the fact that differential Laplacian regularizer can effectively alleviate the unsatisfactory reconstruction results.

\begin{table}[t]\centering
\setlength{\tabcolsep}{2mm}{
  \begin{tabular}{l|c|c|c|c}
    \hline
    \multicolumn{1}{l|}{\multirow{2}{*}{}} & \multicolumn{2}{c|}{$d_C$} & \multicolumn{2}{c}{$d_H$} \\
    \cline{2-5}
     & $Mean$ & $Median$ & $Mean$ & $Median$ \\
    \hline
    SAL~\cite{atzmon2020sal}    & 0.019   & 0.016  & 0.094  & 0.050 \\
    \hline
    SALD~\cite{atzmon2020sald}   & 0.016   & 0.015  & 0.053  & 0.042 \\
    \hline
    IGR~\cite{gropp2020implicit}   & 0.028 & 0.011 & 0.111 &  0.034 \\
    \hline\hline
    Our ($L_{laplace}$)  & 0.017  & 0.009 & 0.068 &  0.026 \\ 
    \hline
    Our ($L_{laplace}+DES$) & \textbf{0.007}  & \textbf{0.007} & \textbf{0.021} & \textbf{0.021}\\
    \hline
  \end{tabular}}
  \caption{
  A quantitative comparison of our method and ablation against IGR~\cite{gropp2020implicit}, SAL~\cite{atzmon2020sal} and SALD ~\cite{atzmon2020sald} on multi-task evaluation dataset.
  }
\label{tab:compare_with_all_methods}\vspace{-5pt}
\end{table}

\subsection{Edge recognition}
\label{subsec:Edge_detection}
Specifically, for each point $p$ in the input point set, we may quickly determine whether it is an edge point according to its differentiable laplace operator $\triangle{f_\theta(x)}$ . Essentially, laplace operator is mainly used to describe the rate of change of gradient, and is often used for edge detection in image processing. From the perspective of differential geometry, it is used to describe the change rate of spatial position normal. Therefore, the larger the laplace operator of the point, the stronger the possibility that the point is an edge point. We threshold the laplace operator $|\triangle{f_\theta(x)}|>\tau$ to obtain a corresponding set of non-edge points $X_{edge}$. We set the parameter $\tau = 20$ throughout our experiments,  as shown in Figure~\ref{fig:edge}.

In addition to IGR~\cite{gropp2020implicit}, we also choose two representative classical non-learning based methods: Voronoi Covariance Measure (VCM)~\cite{merigot2010voronoi}, and  Edge-Aware Resampling (EAR) ~\cite{huang2013edge}, as both have been adopted in the point-set processing routines of the well known CGAL library. 
As reported in Table~\ref{tab:Edge_recognition}, our method completely outperforms these classical methods,
This is mainly because we use the differentiable Laplacian operator of each sampling point as the metric, which can be approximate to the average curvature in the implicit surface representation.
Note that, there are a large number of high-quality edge detection methods based on data-driven. 
We do not use these methods as references here, mainly because ours is a self-supervised learning approach.

\subsection{Normal estimation}
\label{subsec:Normal_estimation}

Essentially, an implicitly represented MLP with softplus activation funtion represents a differentiable Signed Distance Functions $d = f_{\theta}(x)$.
According to the properties of differential geometry, the gradient operator of each point on the implicit surface
$f_{\theta}(x)= 0$ can be regarded as the normal vector of the current point $x$. Therefore, after the training, for each point in the input point cloud, we can directly calculate the gradient operator $\triangledown f_{\theta}(x)$ of the differentiable function $f_{\theta}(x)$ at the current point $x$, that is, the normal vector of the current point $x$. 
The experimental results are reported in Table~\ref{tab:compare_with_all_methods}. The comparison results demonstrate how our method achieves significantly better performance; as immediately quantified by the fact that $d_{angle}$ is larger than the one reported for our method.

\begin{table}[t]\centering
\setlength{\tabcolsep}{2mm}{
  \begin{tabular}{l|c|c|c|c}
    \hline
    \multicolumn{1}{l|}{\multirow{2}{*}{}} & \multicolumn{2}{c|}{$d_C$} & \multicolumn{2}{c}{$d_H$} \\
    \cline{2-5}
               & $Mean$ & $Median$ & $Mean$ & $Median$ \\
    \hline
    $X=0.010$   & 0.0102   & 0.0108   & 0.0509   & 0.0543 \\
    \hline
    $X=0.005$  & 0.0069  & 0.0069	 & 0.0206	& 0.0209 \\
    \hline 
    $X=0.000$   & 0.0055   & 0.0057   & 0.0148   & 0.0153 \\
    \hline\hline
    $D=4,096$  & 0.0075  & 0.0075	 & 0.0350	& 0.0328 \\	
    \hline
    $D=8,192$  & 0.0071  & 0.0072    & 0.0352	& 0.0269 \\		
    \hline
    $D=16,384$ & 0.0069  & 0.0069	 & 0.0206	& 0.0209 \\	
    \hline
  \end{tabular}}
  \caption{
Algorithm performance with respect to noise $X$ and sampling density $D$. 
  }
\label{tab:sample_and_density}\vspace{-5pt}
\end{table}

\subsection{Analysis of parameters and networks}
\textbf{Effect of noise.}
We stress test Laplacian regularizer by increasing the level of noise. Specifically, we randomly add a Gaussian noise whose mean is $0$ and variance is $X$ to each sampling point on the surface of the 3D shape, where we tested four values of $X=\{0, 0.005, 0.01, 0.02 \}$. In each case, the implicit neural surface was trained with the noise-added data. 
Table~\ref{tab:sample_and_density} shows the quantitative results. As we can observe that, the Laplacian regularizer, even when trained with noisy data, can still out-perform these state-of-the-art methods ~\cite{gropp2020implicit,atzmon2020sald,atzmon2020sal} when they are tested on point cloud with 0.005 noise. 

\textbf{Effect of density.}
We also train our method on point clouds at a reduced density. Specifically, for each 3D shape, we sampled a different number 
$D$ of points to verify whether our network could handle the sparser point clouds, where $D=\{4,096, 8,192, 16,384 \}$. (Results in Table~\ref{tab:sample_and_density} reveal a similar trend as from the previous stress test. Namely, our network, when trained on sparser point clouds, can still outperform these state-of-the-art methods ~\cite{gropp2020implicit,atzmon2020sald,atzmon2020sal} when they are tested on or trained on data at full resolution (16,384 points).

\textbf{Effect of $L_{laplacian}$.}
To evaluate the effectiveness of loss $L_{laplacian}$,  
We incorporate this into another state-of-the-art method, SALD \cite{atzmon2020sald},
This qualitative result is shown in Figure~\ref{fig:SALD+LAP+DES}, we can find that, compared with the original algorithm, the reconstruction quality can be effectively improved by incorporating Laplacian.
This is mainly because the differentiable Laplacian regularizer can effectively alleviate the unsatisfactory reconstruction results.



\begin{figure*}[ht]
  \centering
  \includegraphics[width=\textwidth]{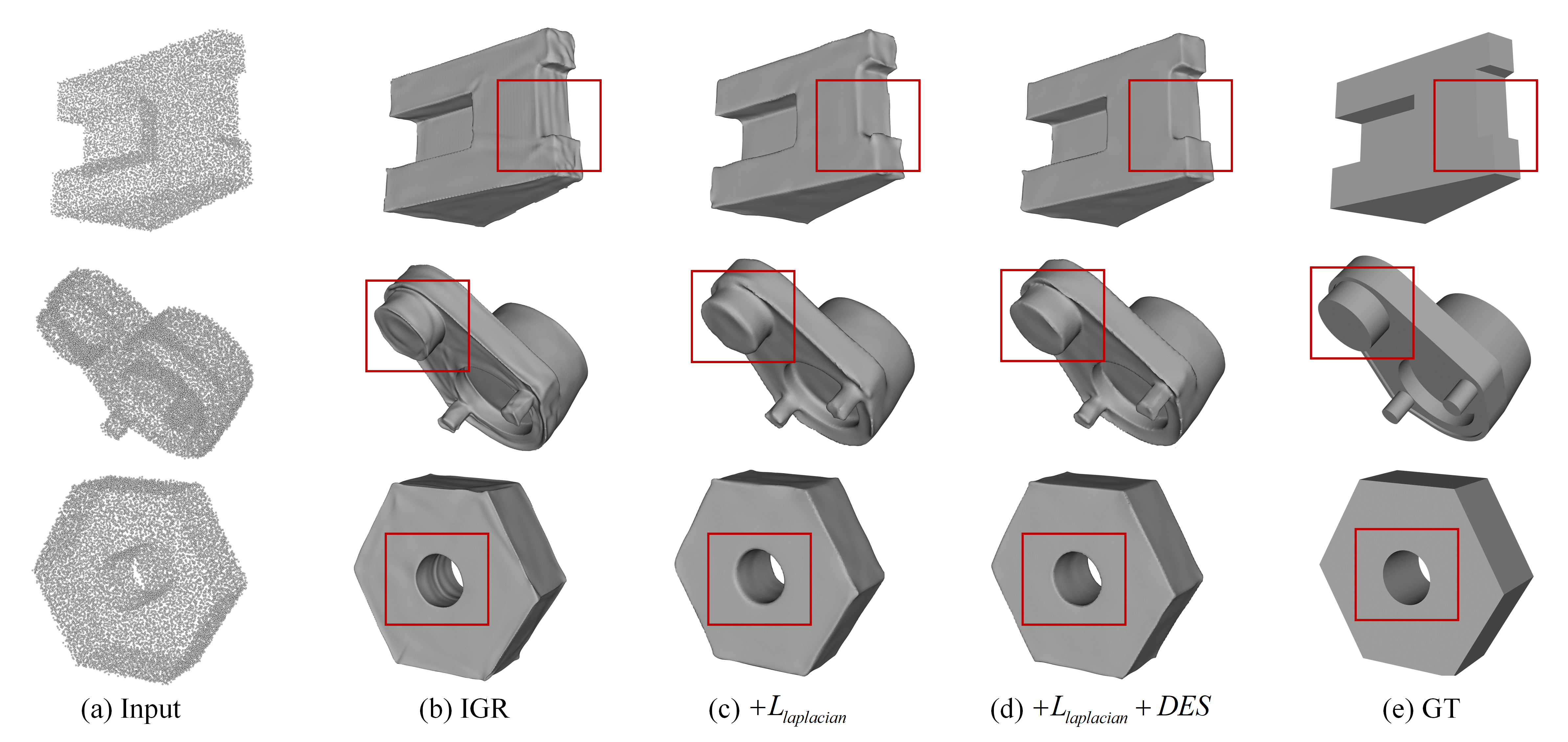}
  \caption{Visualization examples of differential Laplacian regularizer (c) and $dynamic$ $edge$ $sampling$ strategy (d).}
  \label{fig:IGR+L4+DES}
\end{figure*}

\textbf{Dynamic edge sampling.}
We evaluate the effect of $dynamic$ $edge$ $sampling$ strategy on reconstruction quality. 
We experiment with the $dynamic$ $edge$ $sampling$, while keeping all other parameters the same. 
From Table~\ref{tab:compare_with_all_methods} and Figure~\ref{fig:MoreResults_1} and ~\ref{fig:IGR+L4+DES} , we can see that at the sharp edges, we can effectively improve the quality of modeling compared with state-of-the-art methods (Table~\ref{tab:compare_with_all_methods} (rows 1~3)) and the baseline method without $dynamic$ $edge$ $sampling$, this is largely due to the$dynamic$ $edge$ $sampling$ strategy for sampling near the sharp edge of input point cloud, which can effectively avoid the regularizer from smoothing all regions.

\begin{table}[t]\centering
\setlength{\tabcolsep}{3mm}{
  \begin{tabular}{l|c|c|c}
    \hline
     & $d_C$ & $d_H$ & $d_{angle}$ \\
    \hline\hline
    IGR~\cite{gropp2020implicit}   & 0.028 & 0.111 &  0.514 \\
    \hline\hline
    Our ($+L_{laplacian}$)  & 0.017  & 0.068 &  0.274 \\
    \hline
    Our ($+L_{laplacian}+DES$) & \textbf{0.009}  & \textbf{0.036} & \textbf{0.133}\\
    \hline
  \end{tabular}}
  \caption{
Ablation studies – We evaluate the quantitative performance of our method with/without components $L_{laplacian}$ and $dynamic$ $edge$ $sampling$ (DES).
  }
\label{tab:IGR_LAP_DES}\vspace{-5pt}
\end{table}

\begin{table}[t]\centering
\setlength{\tabcolsep}{2.5mm}{
  \begin{tabular}{l|c|c|c|c}
    \hline
     & ECD & IoU & Precision & Recall \\
    \hline\hline
    VCM~\cite{merigot2010voronoi}    & 0.0017 & 0.1925 & 0.2238 & 0.5998 \\
    \hline
    EAR~\cite{huang2013edge}         & 0.0071 & 0.1146 & 0.2399 &  0.1933\\
    \hline\hline
    IGR~\cite{gropp2020implicit}     & 0.0063  & 0.0880  & 0.0958 & 0.5620\\
    \hline\hline
    Our                              & \textbf{0.0015} & \textbf{0.2375} & \textbf{0.2665}  & \textbf{0.6934}\\
    \hline
  \end{tabular}}
\caption{Comparison state-of-the-art edge recognition techniques - VCM~\cite{merigot2010voronoi}, EAR~\cite{huang2013edge}, and IGR~\cite{gropp2020implicit}.}
\label{tab:Edge_recognition}\vspace{-5pt}
\end{table}

\pdfoutput=1
\pdfoutput=1
\pdfoutput=1
\pdfoutput=1
\pdfoutput=1
\section{Conclusion and Limitation}

\begin{figure}[t]
  \centering
  \includegraphics[width=1\linewidth]{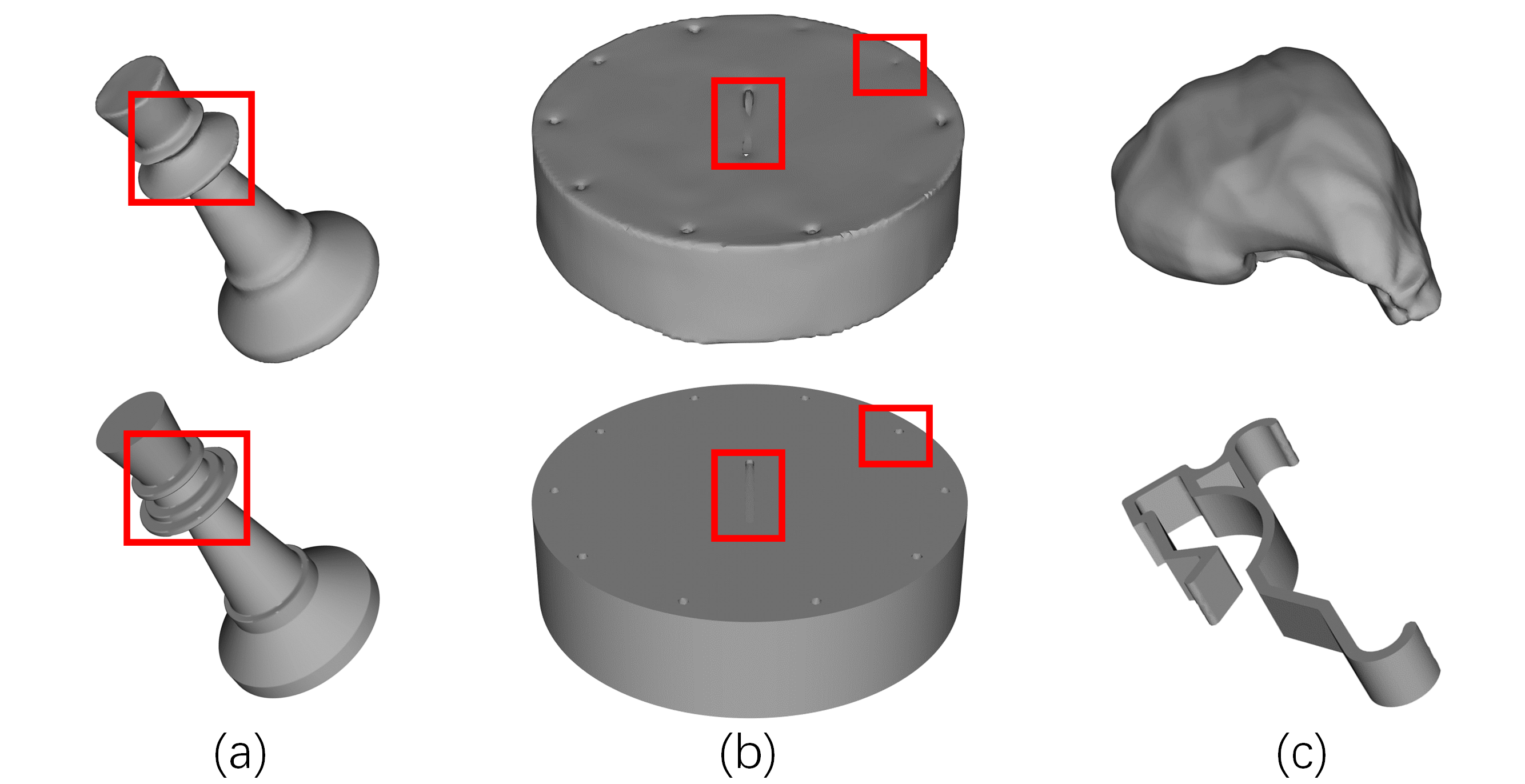}
  \caption{Failure cases.}
  \label{fig:failure_cases}
  \vspace{-8pt}
\end{figure}

We present a differential Laplacian regularizer for neural implicit representation directly from input point cloud without normal supervision. 
More specifically, we use the infinite differentiability property of implicit neural representation
to propose a differentiable Laplacian regularizer, which can effectively alleviate the unsatisfactory reconstruction results. 
Meanwhile, we propose a $dynamic$ $edge$ $sampling$ strategy for sampling near the sharp edge of input point cloud, which can effectively avoid the Laplacian regularizer from smoothing all regions, so as to effectively improve the quality
of reconstruction results while maintaining the edge.
Moreover, the differentiable Laplacian regularizer can be easily and intuitively incorporated into implicit neural surface representations. 
We carefully evaluate its generation quality through a series of ablation studies, which show that our method significantly improve the quality of 3D reconstruction.
In addition to 3D reconstruction, our method can also be conveniently applied to other point cloud analysis tasks, including edge extraction and normal estimation, etc.

\textbf{Limitation.}
Our approach has a few limitations, which point out the directions of future study. 
Some representative failure cases are shown in Figure~\ref{fig:failure_cases}.
First, our method is prone to problems in the reconstruction of ultra-thin geometric structures, probably because the point cloud data is noisy, resulting in the geometric structure has been completely destroyed.
Second, Our method for extremely detailed structure may be overlooked, resulting in incorrect reconstruction results.

\section*{Acknowledgement}
We thank the anonymous reviewers for their valuable comments. This work was supported in part by Natural Science Foundation of China (62102328),
and Fundamental Research Funds for the Central Universities (SWU120076).

\ifCLASSOPTIONcaptionsoff
  \newpage
\fi



\bibliographystyle{IEEEtran}
\bibliography{egbib}

%




\end{document}